\documentclass[runningheads]{llncs}
\usepackage{epstopdf}
\usepackage{tabularx,threeparttable}
\usepackage{array}
\usepackage[hidelinks]{hyperref}
\usepackage{graphicx}
\usepackage{subfigure}
\usepackage{amssymb, amsmath, bm}
\usepackage{mathtools}
\usepackage{mathrsfs}
\usepackage{booktabs}
\usepackage{multirow}
\usepackage{cite}
\usepackage{url}
\usepackage{color}
\usepackage{dsfont}

\usepackage{multirow}

\usepackage{array}
\newcommand{\PreserveBackslash}[1]{\let\temp=\\#1\let\\=\temp}
\newcolumntype{C}[1]{>{\PreserveBackslash\centering}p{#1}}

\def\ie{\emph{i.e.}}
\def\eg{\emph{e.g.}}
\def\etal{{\em et al.}}

\newcolumntype{?}[1]{!{\vrule width #1}}
\usepackage{bbding}

\begin{document}
	
\title{Dual-Teacher: Integrating Intra-domain and Inter-domain Teachers for Annotation-efficient Cardiac Segmentation}
    \author{Kang Li\inst{1} 
    \and Shujun Wang\inst{1} 
    \and Lequan Yu\inst{2}\Envelope 
    \and Pheng-Ann Heng\inst{1,3} 
    }
    \authorrunning{K. Li et al.}
	\institute{Dept. of Computer Science and Engineering,
	The Chinese University of Hong Kong
    \email{ \{kli, sjwang, pheng\}@cse.cuhk.edu.hk}
	\and Dept. of Radiation Oncology, Stanford University \\
    \email{ylqzd2011@gmail.com}
	\and Guangdong Provincial Key Laboratory of Computer Vision and Virtual Reality Technology, Shenzhen Institutes of Advanced Technology,\\ Chinese Academy of Sciences, Shenzhen, China
	}
	
\maketitle              

\pagestyle{headings}
\setcounter{page}{1}


\begin{abstract}
    Medical image annotations are prohibitively time-consuming and expensive to obtain.
    To alleviate annotation scarcity, many approaches have been developed to efficiently utilize extra information, \eg, semi-supervised learning further exploring plentiful unlabeled data, domain adaptation including multi-modality learning and unsupervised domain adaptation resorting to the prior knowledge from additional modality.
    In this paper, we aim to investigate the feasibility of simultaneously leveraging abundant unlabeled data and well-established cross-modality data for annotation-efficient medical image segmentation.
    To this end, we propose a novel semi-supervised domain adaptation approach, namely Dual-Teacher,
    where the student model not only learns from labeled target data (\eg, CT), but also explores unlabeled target data and labeled source data (\eg, MR) by two teacher models.
    Specifically, the student model learns the knowledge of unlabeled target data from intra-domain teacher by encouraging prediction consistency, as well as the shape priors embedded in labeled source data from inter-domain teacher via knowledge distillation.
    Consequently, the student model can effectively exploit the information from all three data resources and comprehensively integrate them to achieve improved performance.
    We conduct extensive experiments on MM-WHS 2017 dataset and demonstrate that our approach is able to concurrently utilize unlabeled data and cross-modality data with superior performance, outperforming semi-supervised learning and domain adaptation methods with a large margin.

\end{abstract}


\section{Introduction}
Deep convolutional neural networks (CNNs) have made great progress in various medical image segmentation applications~\cite{milletari2016v, ronneberger2015unet}. The success is partially relied on massive datasets with abundant annotations. However, collecting and labeling such large-scaled dataset is prohibitively time-consuming and expensive, especially in medical area, since it requires diagnostic expertise and meticulous work~\cite{litjens2017survey}.
Plenty of efforts have been devoted to alleviate annotation scarcity by utilizing extra supervision.
Among them, semi-supervised learning and domain adaptation are two widely studied learning approaches and increasingly gain people's interests.

Semi-supervised learning (SSL) aims to leverage unlabeled data to reduce the usage of manual annotations~\cite{lee2013pseudo,laine2016temporal,tarvainen2017mean}.
For example, Lee~\etal~\cite{lee2013pseudo} proposed to generate the pseudo labels of unlabeled data by a pretrained model, and utilize them to further finetune the training model for performance improvements.
Recently, self-ensembling methods~\cite{laine2016temporal,tarvainen2017mean} have achieved state-of-the-art performance in many semi-supervised learning benchmarks.
Laine~\etal~\cite{laine2016temporal} proposed the temporal ensembling method to encourage the consensus between the exponential moving average (EMA) predictions and current predictions for unlabeled data.
Tarvainen~\etal~\cite{tarvainen2017mean} proposed the mean-teacher framework to force prediction consistency between current training model and the corresponding EMA model.
Although semi-supervised learning has made great progress on utilizing the unlabeled data within the same domain, it leaves rich cross-modality data unexploited.
Considering that multi-modality data is widely available in medical imaging field, recent works have studied on domain adaptation (DA) to leverage the shape priors of another modality for enhanced segmentation performance~\cite{ghafoorian2017transfer,perone2019unsupervised,jiang2018tumor,huo2018synseg}
Among them, multi-modality learning (MML) exploits the labeled data from a related modality (\ie, source domain) to facilitate the segmentation on the modality of interest (\ie, target domain)~\cite{van2018learning, valindria2018multi,jue2019integrating,dou2020unpaired}.
Valindria~\etal~\cite{valindria2018multi} proposed a dual-stream approach to integrate the prior knowledge from unpaired multi-modality data for improved multi-organ segmentation, and suggested X-shape achieving the leading performance among all architectures.
Since multi-modality learning requires annotations on two modality data, unsupervised domain adaptation (UDA) extends it with a broader application potential~\cite{orbes2019knowledge, dou2018unsupervised,chen2019synergistic}.
In UDA setting, source domain annotations are still required, while none target domain annotation is needed.
Contemporary unsupervised domain adaptation methods attempt to extract domain-invariant representations, where Dou~\etal~\cite{dou2018unsupervised} investigated in feature space and Chen~\etal~\cite{chen2019synergistic} explored both feature-level and image-level in a synergistic manner.

All approaches mentioned above have exhibited their feasibility in medical area.
However, semi-supervised learning simply concentrates on leveraging the unlabeled data affiliated to the same domain as labeled ones, ignoring the rich prior knowledge (\eg, shape priors) cross modalities.
While domain adaptation can utilize cross-modality prior knowledge, it still has considerable space for improvement.
These motivate us to explore the feasibility of integrating the merits of both semi-supervised learning and domain adaptation by \emph{concurrently leveraging all available data resources}, including limited labeled target data, abundant unlabeled target data and well-established labeled source data, to enhance the segmentation performance on target domain.

\begin{figure*}[t]
\centering
\includegraphics[width=0.82\textwidth]{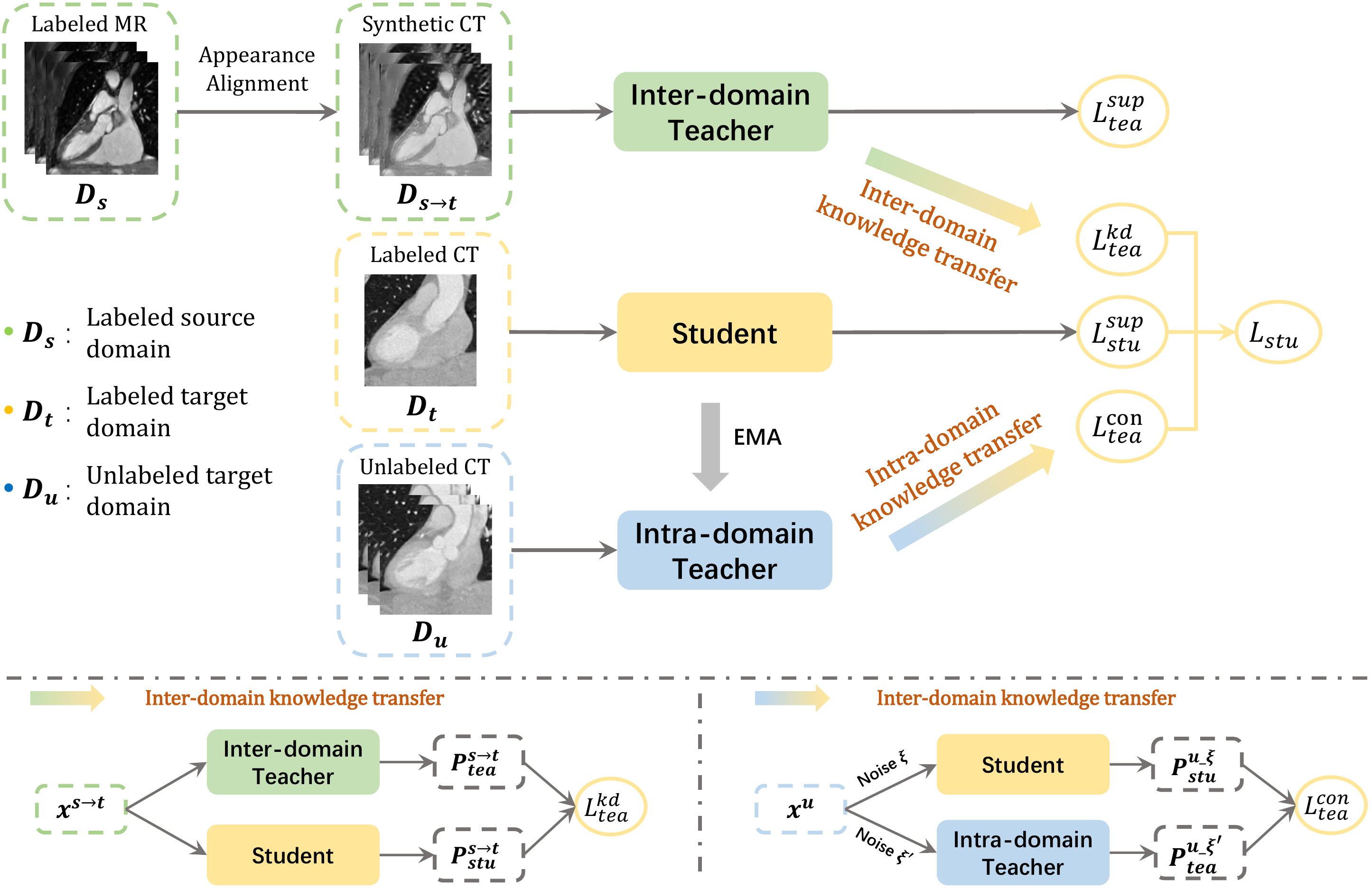}
\caption{Overview of our framework. The student model learns from $\mathcal{D}_{t}$ by the $L_{stu}^{sup}$ loss, and concurrently acquires the knowledge of $\mathcal{D}_{s}$ from inter-domain teacher by knowledge distillation loss $L_{stu}^{kd}$, as well as the knowledge of $\mathcal{D}_{u}$ from intra-domain teacher by the consistency loss $L_{stu}^{con}$. In this way, the student model would integrate and leverage knowledge of $\mathcal{D}_{s}$, $\mathcal{D}_{t}$ and $\mathcal{D}_{u}$ simultaneously, leading to better generalization on target domain. In the inference phase, only the student model is used to predict.}
\label{fig:framework}
\end{figure*}

In this paper, we propose a novel semi-supervised domain adaptation framework, namely Dual-Teacher, to simultaneously leverage abundant unlabeled data and widely-available cross-modality data to mitigate the need for tedious medical annotations.
We implement it with the teacher-student framework~\cite{liu2016dual} and adopt two teacher models in the network training, where one teacher guides the student model with intra-domain knowledge embedded in unlabeled target domain (\eg, CT), while another teacher instructs the student model with inter-domain knowledge beneath labeled source domain (\eg, MR).
To be specific, our Dual-Teacher framework consists of three components:
(1) intra-domain teacher, which employs the self-ensembling model of the student network to leverage unlabeled target data and transfers the acquired knowledge to student model by forcing prediction consistency;
(2) inter-domain teacher, which adopts an image translation model, \ie, CycleGAN~\cite{zhu2017unpaired}, to narrow the appearance gap cross modalities and transfers the prior knowledge in the source domain to student model via knowledge distillation;
and (3) student model, which not only directly learns from limited labeled target data, but also grasps auxiliary intra-domain and inter-domain knowledge transferred from two teachers.
Our whole framework is trained in an end-to-end manner to seamlessly integrate the knowledge of all data resources into the student model.
We extensively evaluated our approach on MM-WHS 2017 dataset~\cite{zhuang2019evaluation}, and achieved superior performance compared to semi-supervised learning methods and domain adaptation methods.
%

\section{Methodology}
In our problem setting, we are given a set of source images and their annotations in source domain (\eg, labeled MR data) as  $\mathcal{D}_{s}=\left\{\left(\mathbf{x}_{i}^{s}, y_{i}^{s}\right)\right\}_{i=1}^{m_{s}}$.
In addition, we are also given a limited number of annotated target domain samples (\eg, labeled CT data) as $\mathcal{D}_{t}=\left\{\left(\mathbf{x}_{i}^{t}, y_{i}^{t}\right)\right\}_{i=1}^{m_{t}}$, and abundant unlabeled  target domain data (\eg, unlabeled CT data) as $\mathcal{D}_{u}=\left\{\left(\mathbf{x}_{i}^{u}\right)\right\}_{i=1}^{m_{u}}$.
Normally, we assume $m_{t}$ is far less than $m_{u}$.
Our goal is to exploit $\mathcal{D}_{s},~ \mathcal{D}_{t}$~and $\mathcal{D}_{u}$~to enhance the performance in target domain (\eg, CT).
Fig.~\ref{fig:framework} overviews our proposed Dual-Teacher framework, which consists of an inter-domain teacher model, an intra-domain teacher model, and a student model.
The inter-domain teacher model and intra-domain teacher model explore the knowledge beneath $\mathcal{D}_{s}$ and $\mathcal{D}_{u}$, respectively, and simultaneously transfer the knowledge to the student model for comprehensive integration and thorough exploitation.

\subsection{Inter-domain Teacher}
Despite the consistent shape priors shared between source domain (\eg, MR) and target domain (\eg, CT), they are distinct in many aspects like appearance and image distribution~\cite{pan2009survey, hoffman2018cycada}. Considering that, we attempt to reduce the appearance discrepancy first by using an appearance alignment module.
Various image translation models can be adopted. Here we use CycleGAN~\cite{zhu2017unpaired} to translate source samples $x^{s}$ to synthetic target-style samples $x^{s \rightarrow t}$ for synthetic target set $\mathcal{D}_{s\rightarrow t}$.
After appearance alignment, we input synthetic samples $x^{s \rightarrow t}$ into the inter-domain teacher, which is implemented as a segmentation network. With the supervision of corresponding labels $y^{s}$, the inter-domain teacher is able to learn the prior knowledge in source domain by $\mathcal{L}_{tea}^{seg}$ following
\begin{equation}
    \mathcal{L}_{tea}^{seg}=\mathcal{L}_{\mathrm{ce}}\left(y^{s}, p_{tea}^{s\rightarrow t}\right)+\mathcal{L}_{\mathrm{dice}}\left(y^{s}, p_{tea}^{s\rightarrow t}\right),
\end{equation}
where $\mathcal{L}_{\mathrm{ce}}$ and $\mathcal{L}_{\mathrm{dice}}$ denote cross-entropy loss and dice loss, respectively, and $p_{tea}^{s\rightarrow t}$ represents the inter-domain teacher predictions taking $x^{s\rightarrow t}$ as inputs.
To transfer the acquired knowledge from inter-domain teacher to the student, we further feed the same synthetic samples $x^{s \rightarrow t}$ into both inter-domain teacher model and student model.
Since the inter-domain teacher has acquired reliable source domain knowledge from its annotations, we encourage the student model to produce similar outputs as inter-domain teacher model via knowledge distillation loss $\mathcal{L}_{tea}^{kd}$.
Following previous works~\cite{hinton2015distilling,anil2018large}, we formulate $\mathcal{L}_{tea}^{kd}$ as
\begin{equation}
\mathcal{L}_{\mathrm{tea}}^{kd}=\mathcal{L}_{\mathrm{ce}}\left(p_{tea}^{s \rightarrow t}, p_{stu}^{s \rightarrow t}\right),
\label{eq: know_distill}
\end{equation}
where $p_{tea}^{s \rightarrow t}$ and $p_{stu}^{s \rightarrow t}$ represent the predictions of inter-domain teacher model and student model, respectively.

\subsection{Intra-domain Teacher}
As $\mathcal{D}_{u}$ has no expert-annotated labels to directly guide network learning, recent works~\cite{tarvainen2017mean} propose to temporally ensemble the models in different training steps for reliable predictions.
Inspired by them, we design the intra-domain teacher model following the same network architecture as student model and its weights $\theta^{\prime}$ are updated as the exponential moving average (EMA) of the student model weights $\theta$ in different training steps.
Specifically, at training step $t$, the weights of intra-domain teacher model $\theta_{t}^{\prime}$ are updated as
\begin{equation}
\theta_{t}^{\prime}=\alpha \theta_{t-1}^{\prime}+(1-\alpha) \theta_{t},
\end{equation}
where $\alpha$ is the EMA decay rate to control updating rate.
To transfer the knowledge from intra-domain teacher to the student, we add different noise $\xi$ and $\xi^{\prime}$ to the same unlabeled sample and feed them into intra-domain teacher model and student model, respectively.
Given small perturbation operations, \eg, Gaussian noise, the outputs between the student model and the corresponding EMA model (\ie, the intra-domain teacher model) should be the same.
Therefore, we encourage them to generate consistent predictions via consistency loss $\mathcal{L}_{\mathrm{tea}}^{con}$ as
\begin{equation}
    \mathcal{L}_{tea}^{con}=\mathcal{L}_{\mathrm{mse}}\left(f\left(x^{u};\theta,\xi\right), f\left(x^{u};\theta^{\prime},\xi^{\prime}\right)\right),
    \label{eq: mean-tea}
\end{equation}
where $\mathcal{L}_{\mathrm{mse}}$ denotes the mean squared error loss.  $f\left(x^{u};\theta,\xi\right)$ and $f\left(x^{u};\theta_{t},\xi^{\prime}\right)$ represent the outputs of the student model (with weight $\theta$ and noise $\xi$) and intra-domain teacher model (with weight $\theta^{\prime}$ and noise $\xi^{\prime}$), respectively.

\subsection{Student Model and Overall Training Strategies}
For the student model, it explicitly learns from $\mathcal{D}_{t}$ with the supervision of its labels via the segmentation loss $\mathcal{L}_{stu}^{seg}$. Meanwhile, it also concurrently acquires the knowledge of $\mathcal{D}_{s}$ and $\mathcal{D}_{u}$ from two teacher models and comprehensively integrates them as a united cohort.
In particular, the student model attains inter-domain knowledge by knowledge distillation loss $\mathcal{L}_{tea}^{kd}$ as Eq.~\eqref{eq: know_distill}, and intra-domain knowledge by prediction consistency loss $\mathcal{L}_{tea}^{con}$ as Eq.~\eqref{eq: mean-tea}.
Overall, the training objective for the student model is formulated as
\begin{equation}
\begin{aligned}
\mathcal{L}_{stu}^{seg}&=\mathcal{L}_{\mathrm{ce}}\left(y^{t}, p_{stu}^{t}\right)+\mathcal{L}_{\mathrm{dice}}\left(y^{t}, p_{stu}^{t}\right)\\
\mathcal{L}_{stu}&=\mathcal{L}_{stu}^{seg} + \lambda_{kd}\mathcal{L}_{tea}^{kd} + \lambda_{con}\mathcal{L}_{tea}^{con},
\end{aligned}
\end{equation}
where $\lambda_{kd}$ and $\lambda_{con}$ are hyperparameters to balance the weight of $\mathcal{L}_{tea}^{kd}$ and $\mathcal{L}_{tea}^{con}$.

Our whole framework is updated in an end-to-end manner. We first optimize the inter-domain teacher model, then update the intra-domain teacher model with the EMA parameters of the student network, and optimize the student model in the last.
In this way, no pre-training stage would be required and the student model updates its parameters synchronously along with teacher models in an online manner.

\section{Experiments}

\subsubsection{Dataset and pre-processing}
We evaluated our method in Multi-modality Whole Heart Segmentation (MM-WHS) 2017 dataset~\cite{zhuang2019evaluation}, which provided 20 annotated MR and 20 annotated CT volumes.
We employed CT as target domain and MR as source domain, and randomly split 20 CT volumes into four folds to perform four-fold cross validation. In each fold, we validated on five CT volumes, and took 20 MR volumes as $\mathcal{D}_{s}$, five randomly chosen CT volumes as $\mathcal{D}_{t}$ and the remaining 10 CT volumes as $\mathcal{D}_{u}$ to train our framework.
For pre-processing, we resampled all data with unit spacing and cropped them into $256\times256$ centering at the heart region, following previous work~\cite{chen2019synergistic}. To avoid overfitting, we applied on-the-fly data augmentation with random affine transformations and random rotation.
We evaluated our method with dice coefficient on all seven heart substructures, including the left ventricle blood cavity (LV), the right ventricle blood cavity (RV), the left atrium blood cavity (LA), the right atrium blood cavity (RA), the myocardium of the left ventricle (MYO), the ascending aeorta (AA), and the pulmonary artery (PA)~\cite{zhuang2019evaluation}.

\subsubsection{Implementation details}
In our framework, the student model and two teacher models were implemented with the same network backbone, U-Net~\cite{ronneberger2015unet}.
We empirically set $\lambda_{kd}$ as $0.1$ for inter-domain teacher.
For intra-domain teacher, we closely followed the experiment configurations in previous work~\cite{yu2019uncertainty}, where the EMA decay rate $\alpha$ was set to $0.99$ and the hyperparameter $\lambda_{con}$ was dynamically changed over time with the function $\lambda_{con}(t)=0.1 * e^{\left(-5\left(1-t / t_{\max }\right)^{2}\right)}$, where $t$ and $t_{max}$ denote the current and the last training epoch respectively and $t_{max}$ is set to $50$.
To optimize the appearance alignment module, we followed the setting in~\cite{zhu2017unpaired} and used Adam optimizer with learning rate $0.0001$ to optimize the student model and two teacher models until the network converge.
\begin{table*}[t]
\centering
\caption{Comparison with other methods. The dice of all heart substructures and the average of them are reported here.}
\resizebox{1.0\textwidth}{!}{
    \begin{tabular}{c|c|C{1.2cm}|c|c|c|c|c|c|c}
    \toprule[1pt]
    \multicolumn{2}{c|}{\multirow{2}{*}{Method}} & \multirow{2}{*}{Avg} & \multicolumn{7}{c}{Dice of heart substructures} \\ \cline{4-10}
    \multicolumn{2}{c|}{} & &MYO & LA & LV & RA & RV & AA & PA\\
    \midrule
    \multicolumn{2}{c|}{Supervised-only ($\mathcal{D}_{t}$)} & 0.7273 & 0.7113 & 0.7346 & 0.8086 & 0.7099 & 0.6524 & 0.8707 & 0.6037 \\ \hline

    \multirow{2}{*}{\begin{tabular}[c]{@{}c@{}}UDA\\ $(\mathcal{D}_{s}, \mathcal{D}_{u})$\end{tabular}} & Dou~\etal\cite{dou2018unsupervised} & 0.6635 & 0.5664 & 0.7655 & 0.7654 & 0.6230 &  0.6600 & 0.7138 & 0.5505 \\
      & Chen~\etal\cite{chen2019synergistic} & 0.7138 & 0.6573 & 0.8290 &  0.8306 & 0.7804 & 0.7082 & 0.7089 & 0.4827 \\ \hline

    \multirow{3}{*}{\begin{tabular}[c]{@{}c@{}}MML\\ $(\mathcal{D}_{s}, \mathcal{D}_{t})$\end{tabular} } & Finetune & 0.7313 & 0.7533 & 0.8081 & 0.7825  & 0.6412 & 0.5928 & 0.8466 & 0.6943  \\
                          & Joint training & 0.7875 & 0.7816 & 0.8312 & 0.8469 & 0.7699 & 0.7008 & 0.8802 & 0.7019\\
                          & X-shape~\cite{valindria2018multi} & 0.7643 & 0.7317 & 0.8361 & 0.8432 & 0.7259 & 0.7453 & 0.8968 & 0.5709\\ \hline

    \begin{tabular}[c]{@{}c@{}}SSL\\ $(\mathcal{D}_{u}, \mathcal{D}_{t})$\end{tabular} & MT\cite{tarvainen2017mean} & 0.8165 & 0.7764 & 0.8712 & 0.8748 & 0.7930 & 0.7051 & 0.9274 & 0.7677 \\ \hline

    \begin{tabular}[c]{@{}c@{}}SSDA\\ $(\mathcal{D}_{s}, \mathcal{D}_{u}, \mathcal{D}_{t})$\end{tabular} & \textbf{Ours} & \bm{$0.8604$} & \bm{$0.8143$} & \bm{$0.8784$} & \bm{$0.9054$} & \bm{$0.8449$} & \bm{$0.8342$} & \bm{$0.9412$} & \bm{$0.8043$} \\

    \bottomrule[1pt]
    \end{tabular}
}
\label{tab: main}
\end{table*}

\begin{figure*}[t]
\centering
\includegraphics[width=0.98\textwidth]{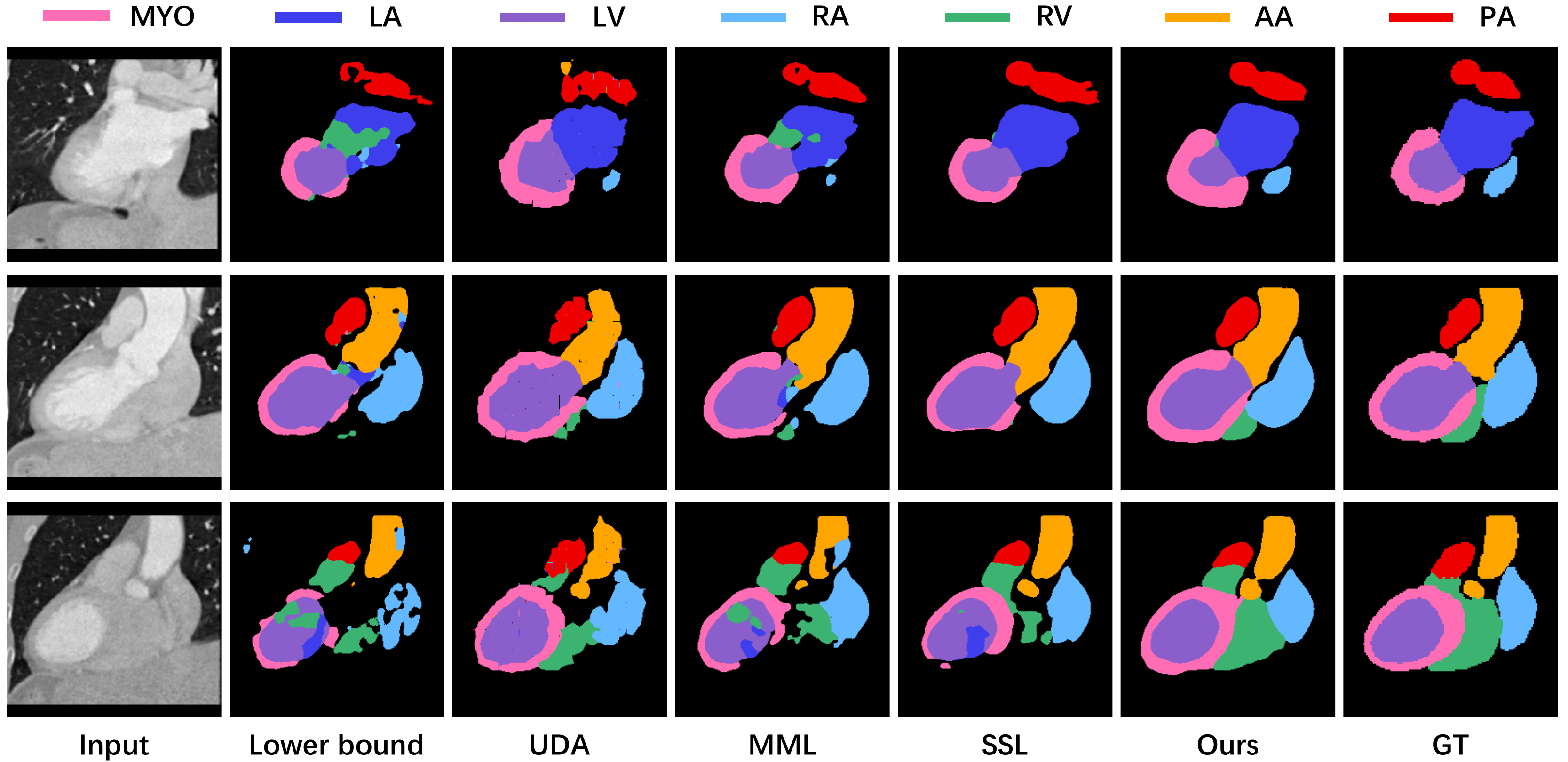}
\caption{Visual comparisons with other methods. Due to page limit, we only present the methods with best mean dice in MML and UDA (\ie, Joint-training and Chen~\etal~\cite{chen2019synergistic}). As observed, our predictions are more similar to the ground truth than others.}
\label{fig:qualitative}
\end{figure*}

\subsubsection{Comparison with other methods}
To demonstrate the effectiveness of our proposed semi-supervised domain adaptation method (SSDA) for leveraging unlabeled data and cross-modality data, we compare with both semi-supervised learning methods and domain adaptation methods.
We first compare with the model trained with only limited labeled CT data $\mathcal{D}_{t}$ (referred as Supervised-only), and take mean-teacher (MT) method~\cite{tarvainen2017mean} in semi-supervised learning (SSL) for comparison.
For domain adaptation methods, besides straightforward methods like finetune and joint training, we also compare with X-shape model~\cite{valindria2018multi} in multi-modality learning (MML).
Meanwhile, we consider two unsupervised domain adaptation methods (UDA)  for comparisons, \ie, Dou~\etal~\cite{dou2018unsupervised} and Chen~\etal~\cite{chen2019synergistic}, which achieve the state-of-the-art performance in cardiac segmentation.

As presented in Table~\ref{tab: main}, the supervised-only model achieves $72.73\%$ in mean dice by taking only limited labeled target data $\mathcal{D}_{t}$ in network training.
When two types of data resources are available, UDA methods achieve comparable mean dice to the supervised-only model by utilizing $\mathcal{D}_{s}$ and $\mathcal{D}_{u}$.
Compared with supervised-only method, MML-based Joint training and SSL-based MT~\cite{tarvainen2017mean} methods further improve the segmentation performance with $6.02\%$ and $8.92\%$ in mean dice, respectively, demonstrating the effectiveness of leveraging cross-modality data $\mathcal{D}_{s}$ and unlabeled data $\mathcal{D}_{u}$ for improving segmentation performance.
By simultaneously exploiting all of data resources, our Dual-Teacher outperforms the unsupervised domain adaptation, multi-modality learning and semi-supervised learning methods by a large margin,~\ie, $14.66\%$, $7.29\%$ and $4.39\%$ increase in mean dice respectively, validating the feasibility of our proposed semi-supervised domain adaptation approach.

We also present visual comparisons in Fig.~\ref{fig:qualitative}. Due to page limit, we only present the predictions of the methods with best mean dice in MML (\ie, Joint-training) and UDA (\ie, Chen~\etal~\cite{chen2019synergistic}).
It is observed that our method better identifies heart substructures with clean and accurate boundary, and produces less false positive predictions and more similar results to the ground truth compared with other methods.

\begin{table}[t]
\centering
\caption{Analysis of our method. We report the mean dice of all cardiac substructures.}
\begin{tabular}{c|C{4cm}|C{2cm}}
\toprule[1pt]
\multicolumn{2}{c|}{Methods}                             & Mean Dice \\
\midrule
\multirow{2}{*}{No-Teacher}  & Baseline                 & 0.7330    \\
                             & GAN-baseline             & 0.7510    \\ \hline

\multirow{2}{*}{One-Teacher} & W/o inter-domain teacher & 0.8477    \\
                             & W/o intra-domain teacher & 0.7984    \\ \hline

\multicolumn{2}{c|}{\textbf{Dual-Teacher (Ours)}}                 & 0.8604   \\
\bottomrule[1pt]
\end{tabular}
\label{tab:ablation}
\end{table}

\subsubsection{Analysis of our method}
We further compare with other methods, which also utilize all three types of data in SSDA, and analyze the key components of our method in Table~\ref{tab:ablation}.
For $\mathcal{D}_{s}$, $\mathcal{D}_{t}$, $\mathcal{D}_{u}$ in SSDA, one  straightforward method is to train $\mathcal{D}_{s}$ and $\mathcal{D}_{t}$ jointly, and deploy Pseudo-label method~\cite{lee2013pseudo} to utilize $\mathcal{D}_{u}$, which is considered as our Baseline.
A more effective version of baseline (referred as GAN-baseline) is using appearance alignment module (\eg, CycleGAN~\cite{zhu2017unpaired}) on $\mathcal{D}_{s}$ to minimize appearance difference, and then following the previous routine by joint training synthetic target data $\mathcal{D}_{s\rightarrow t}$ along with $\mathcal{D}_{t}$ and applying Pseudo-label method~\cite{lee2013pseudo} for $\mathcal{D}_{u}$.
For the Baseline and GAN-baseline, no teacher-student scheme is applied.
Moreover, we conduct other experiments: (i) without inter-domain teacher, where we substitute it as a joint-training network attached with appearance alignment module to tackle $\mathcal{D}_{s}$ and $\mathcal{D}_{t}$, and (ii) without intra-domain teacher, where we replace it with Pseudo-label method~\cite{lee2013pseudo} to handle $\mathcal{D}_{u}$.

The results are shown in Table~\ref{tab:ablation}.
Without any knowledge transfer from teacher models, neither the knowledge in $\mathcal{D}_{s}$ or that in $\mathcal{D}_{u}$ would be well-exploited.
Since GAN-baseline adopts special treatments to narrow appearance gap, it performs better than the baseline model, but it still has large room for improvement compared to our method.
Without the intra-domain teacher, the pseudo label bias will gradually accumulated and deteriorate the segmentation performance with $6.20\%$ lower than our Dual-Teacher framework in mean dice.
Without the inter-domain teacher, the performance is $1.27\%$ lower than our method in mean dice, indicating that the prior knowledge of $\mathcal{D}_{s}$ are not effectively utilized.
These comparison results show that each teacher model plays a crucial role in our framework and further improvements could be achieved when combining them together.


\section{Conclusion}
We present a novel annotation-efficient semi-supervised domain adaptation framework for multi-modality cardiac segmentation.
Our method integrates the inter-domain teacher model to leverage cross-modality priors from source domain, and the intra-domain teacher model to exploit the knowledge embedded in unlabeled target data.
Both teacher models transfer the learnt knowledge into the student model, thereby seamlessly combining the merits of semi-supervised learning and domain adaptation.
We extensively evaluated our method in MM-WHS 2017 dataset. Our method can simultaneously utilize cross-modality data and unlabeled data, and outperforms state-of-the-art semi-supervised and domain adaptation methods.
\\
\\
\textbf{Acknowledgments.}
The work described in this paper was supported by Key-Area Research and Development Program of Guangdong Province, China under Project No. 2020B010165004, Hong Kong Innovation and Technology Fund under Project No. ITS/426/17FP and ITS/311/18FP and National Natural Science Foundation of China under Project No. U1813204.

\bibliographystyle{splncs04}
\bibliography{ref}

\begin{thebibliography}{10}
\providecommand{\url}[1]{\texttt{#1}}
\providecommand{\urlprefix}{URL }
\providecommand{\doi}[1]{https://doi.org/#1}

\bibitem{anil2018large}
Anil, R., Pereyra, G., Passos, A., Ormandi, R., Dahl, G.E., Hinton, G.E.: Large
  scale distributed neural network training through online distillation. arXiv
  preprint arXiv:1804.03235  (2018)

\bibitem{chen2019synergistic}
Chen, C., Dou, Q., Chen, H., Qin, J., Heng, P.A.: Synergistic image and feature
  adaptation: Towards cross-modality domain adaptation for medical image
  segmentation. In: Proceedings of the AAAI Conference on Artificial
  Intelligence. vol.~33, pp. 865--872 (2019)

\bibitem{dou2020unpaired}
Dou, Q., Liu, Q., Heng, P.A., Glocker, B.: Unpaired multi-modal segmentation
  via knowledge distillation. In: IEEE Transactions on Medical Imaging (2020)

\bibitem{dou2018unsupervised}
Dou, Q., Ouyang, C., Chen, C., Chen, H., Heng, P.A.: Unsupervised
  cross-modality domain adaptation of convnets for biomedical image
  segmentations with adversarial loss. In: Proceedings of the 27th
  International Joint Conference on Artificial Intelligence. pp. 691--697
  (2018)

\bibitem{ghafoorian2017transfer}
Ghafoorian, M., Mehrtash, A., Kapur, T., Karssemeijer, N., Marchiori, E.,
  Pesteie, M., Guttmann, C.R., de~Leeuw, F.E., Tempany, C.M., van Ginneken, B.,
  et~al.: Transfer learning for domain adaptation in mri: Application in brain
  lesion segmentation. In: International conference on medical image computing
  and computer-assisted intervention. pp. 516--524. Springer (2017)

\bibitem{hinton2015distilling}
Hinton, G., Vinyals, O., Dean, J.: Distilling the knowledge in a neural
  network. arXiv preprint arXiv:1503.02531  (2015)

\bibitem{hoffman2018cycada}
Hoffman, J., Tzeng, E., Park, T., Zhu, J.Y., Isola, P., Saenko, K., Efros, A.,
  Darrell, T.: Cycada: Cycle-consistent adversarial domain adaptation. In:
  International Conference on Machine Learning. pp. 1989--1998 (2018)

\bibitem{huo2018synseg}
Huo, Y., Xu, Z., Moon, H., Bao, S., Assad, A., Moyo, T.K., Savona, M.R.,
  Abramson, R.G., Landman, B.A.: Synseg-net: Synthetic segmentation without
  target modality ground truth. IEEE transactions on medical imaging
  \textbf{38}(4),  1016--1025 (2018)

\bibitem{jiang2018tumor}
Jiang, J., Hu, Y.C., Tyagi, N., Zhang, P., Rimner, A., Mageras, G.S., Deasy,
  J.O., Veeraraghavan, H.: Tumor-aware, adversarial domain adaptation from ct
  to mri for lung cancer segmentation. In: International Conference on Medical
  Image Computing and Computer-Assisted Intervention. pp. 777--785. Springer
  (2018)

\bibitem{jue2019integrating}
Jue, J., Jason, H., Neelam, T., Andreas, R., Sean, B.L., Joseph, D.O., Harini,
  V.: Integrating cross-modality hallucinated mri with ct to aid mediastinal
  lung tumor segmentation. In: International Conference on Medical Image
  Computing and Computer-Assisted Intervention. pp. 221--229. Springer (2019)

\bibitem{laine2016temporal}
Laine, S., Aila, T.: Temporal ensembling for semi-supervised learning. arXiv
  preprint arXiv:1610.02242  (2016)

\bibitem{lee2013pseudo}
Lee, D.H.: Pseudo-label: The simple and efficient semi-supervised learning
  method for deep neural networks. In: Workshop on challenges in representation
  learning, ICML. vol.~3, p.~2 (2013)

\bibitem{litjens2017survey}
Litjens, G., Kooi, T., Bejnordi, B.E., Setio, A.A.A., Ciompi, F., Ghafoorian,
  M., Van Der~Laak, J.A., Van~Ginneken, B., S{\'a}nchez, C.I.: A survey on deep
  learning in medical image analysis. Medical image analysis  \textbf{42},
  60--88 (2017)

\bibitem{liu2016dual}
Liu, F., Deng, C., Bi, F., Yang, Y.: Dual teaching: A practical semi-supervised
  wrapper method. arXiv preprint arXiv:1611.03981  (2016)

\bibitem{milletari2016v}
Milletari, F., Navab, N., Ahmadi, S.A.: V-net: Fully convolutional neural
  networks for volumetric medical image segmentation. In: 2016 Fourth
  International Conference on 3D Vision (3DV). pp. 565--571. IEEE (2016)

\bibitem{orbes2019knowledge}
Orbes-Arteainst, M., Cardoso, J., S{\o}rensen, L., Igel, C., Ourselin, S.,
  Modat, M., Nielsen, M., Pai, A.: Knowledge distillation for semi-supervised
  domain adaptation. In: OR 2.0 Context-Aware Operating Theaters and Machine
  Learning in Clinical Neuroimaging, pp. 68--76. Springer (2019)

\bibitem{pan2009survey}
Pan, S.J., Yang, Q.: A survey on transfer learning. IEEE Transactions on
  knowledge and data engineering  \textbf{22}(10),  1345--1359 (2009)

\bibitem{perone2019unsupervised}
Perone, C.S., Ballester, P., Barros, R.C., Cohen-Adad, J.: Unsupervised domain
  adaptation for medical imaging segmentation with self-ensembling. NeuroImage
  \textbf{194},  1--11 (2019)

\bibitem{ronneberger2015unet}
Ronneberger, O., Fischer, P., Brox, T.: U-net: Convolutional networks for
  biomedical image segmentation. In: International Conference on Medical image
  computing and computer-assisted intervention. pp. 234--241. Springer (2015)

\bibitem{tarvainen2017mean}
Tarvainen, A., Valpola, H.: Mean teachers are better role models:
  Weight-averaged consistency targets improve semi-supervised deep learning
  results. In: Advances in neural information processing systems. pp.
  1195--1204 (2017)

\bibitem{valindria2018multi}
Valindria, V.V., Pawlowski, N., Rajchl, M., Lavdas, I., Aboagye, E.O., Rockall,
  A.G., Rueckert, D., Glocker, B.: Multi-modal learning from unpaired images:
  Application to multi-organ segmentation in ct and mri. In: 2018 IEEE Winter
  Conference on Applications of Computer Vision (WACV). pp. 547--556. IEEE
  (2018)

\bibitem{van2018learning}
Van~Tulder, G., de~Bruijne, M.: Learning cross-modality representations from
  multi-modal images. IEEE transactions on medical imaging  \textbf{38}(2),
  638--648 (2018)

\bibitem{yu2019uncertainty}
Yu, L., Wang, S., Li, X., Fu, C.W., Heng, P.A.: Uncertainty-aware
  self-ensembling model for semi-supervised 3d left atrium segmentation. In:
  International Conference on Medical Image Computing and Computer-Assisted
  Intervention. pp. 605--613. Springer (2019)

\bibitem{zhu2017unpaired}
Zhu, J.Y., Park, T., Isola, P., Efros, A.A.: Unpaired image-to-image
  translation using cycle-consistent adversarial networks. In: Proceedings of
  the IEEE international conference on computer vision. pp. 2223--2232 (2017)

\bibitem{zhuang2019evaluation}
Zhuang, X., Li, L., Payer, C., {\v{S}}tern, D., Urschler, M., Heinrich, M.P.,
  Oster, J., Wang, C., Smedby, {\"O}., Bian, C., et~al.: Evaluation of
  algorithms for multi-modality whole heart segmentation: An open-access grand
  challenge. Medical image analysis  \textbf{58},  101537 (2019)

\end{thebibliography}

\end{document}